\documentclass[10pt,twocolumn,letterpaper]{article}
\usepackage[accsupp]{axessibility}
\usepackage{iccv}
\usepackage{times}
\usepackage{epsfig}
\usepackage{graphicx}
\usepackage{amsmath}
\usepackage{amssymb}
\usepackage{authblk}
\usepackage{verbatim}
\usepackage{multirow}
\usepackage{bm}

\usepackage[pagebackref=true,breaklinks=true,letterpaper=true,colorlinks,bookmarks=false]{hyperref}

\usepackage[breaklinks=true,bookmarks=false]{hyperref}

\iccvfinalcopy 

\def\iccvPaperID{****} 

\begin{document}

\title{PR-Net: Preference Reasoning for Personalized Video Highlight Detection}







\makeatletter
\renewcommand\AB@affilsepx{\quad \protect\Affilfont} 
\makeatother

\author[1,2]{Runnan Chen \thanks{Work done during internship at Tencent Youtu Lab.}}
\author[2]{Penghao Zhou}
\author[3]{Wenzhe Wang}
\author[1]{Nenglun Chen}
\author[2]{Pai Peng}
\author[2]{\\ Xing Sun\thanks{Corresponding author}}
\author[1]{Wenping Wang$^{\dagger}$}

\affil[1]{The University of Hong Kong}
\makeatletter 
\renewcommand\AB@affilsepx{\\ \protect\Affilfont}
\makeatother

\affil[2]{Youtu Lab, Tencent}
\makeatletter 
\renewcommand\AB@affilsepx{\\ \protect\Affilfont}
\makeatother
\affil[3]{Zhejiang University}

\renewcommand\Authands{, }

\maketitle

\pagestyle{empty}
\thispagestyle{empty}

\begin{abstract}
Personalized video highlight detection aims to shorten a long video to interesting moments according to a user's preference, which has recently raised the community's attention. Current methods regard the user's history as holistic information to predict the user's preference but negating the inherent diversity of the user's interests, resulting in vague preference representation. In this paper, we propose a simple yet efficient preference reasoning framework (PR-Net) to explicitly take the diverse interests into account for frame-level highlight prediction. Specifically, distinct user-specific preferences for each input query frame are produced, presented as the similarity weighted sum of history highlights to the corresponding query frame. Next, distinct comprehensive preferences are formed by the user-specific preferences and a learnable generic preference for more overall highlight measurement. Lastly, the degree of highlight and non-highlight for each query frame is calculated as semantic similarity to its comprehensive and non-highlight preferences, respectively. Besides, to alleviate the ambiguity due to the incomplete annotation, a new bi-directional contrastive loss is proposed to ensure a compact and differentiable metric space. In this way, our method significantly outperforms state-of-the-art methods with a relative improvement of 12\% in mean accuracy precision.

\end{abstract}

\begin{figure}
  \centerline{\includegraphics[width=0.5\textwidth]{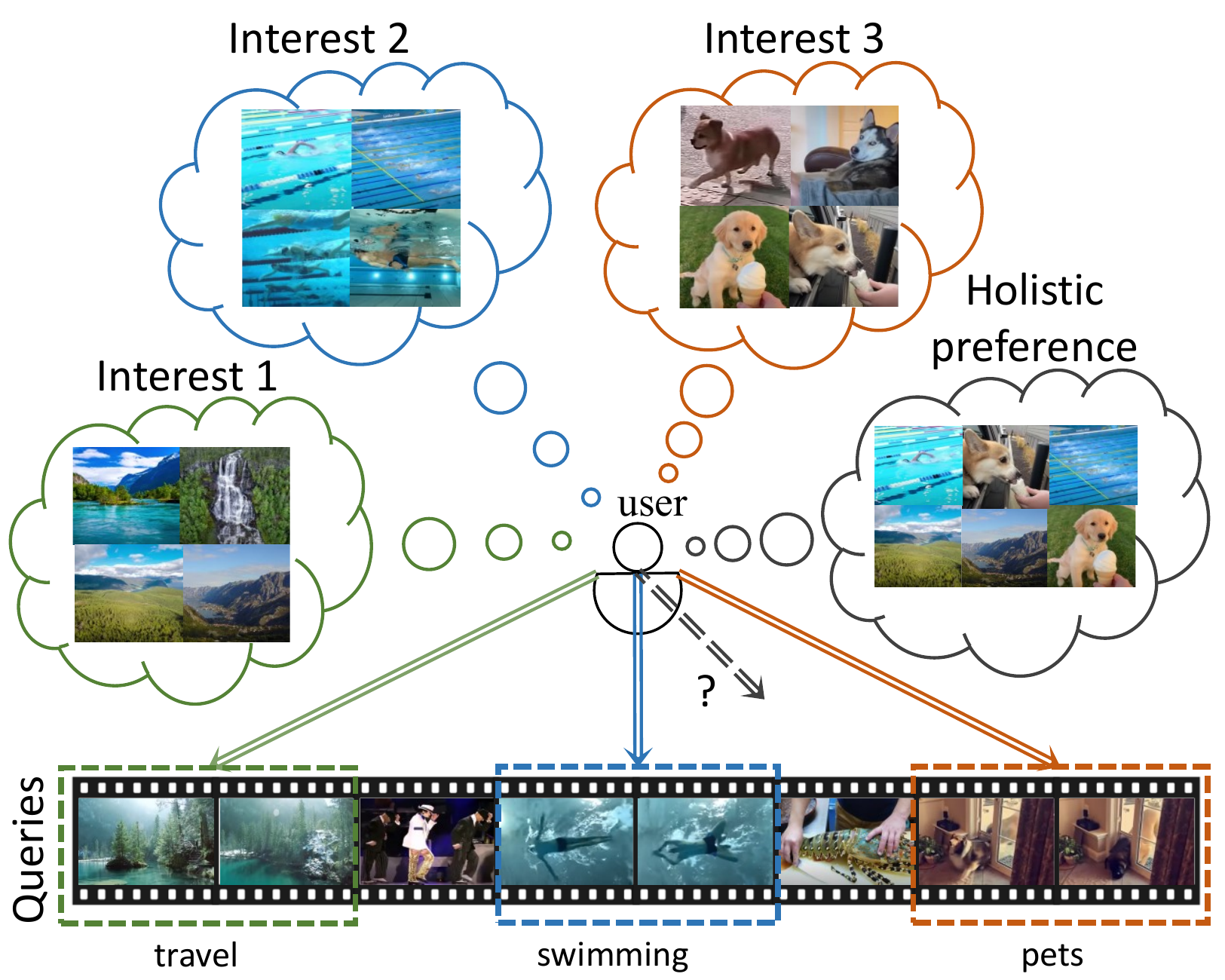}}
  \caption{Considering the diversity of the user's interests is critical for guiding personalized video highlight detection. For example, if a user is interested in travel, swimming and pets, those query frames that are semantically similar to one of the user's interests can be marked as the highlight frames. On the other hand, predicting a holistic user's preference from history will lead to vague preference representation and unsatisfied performance.}
  \label{fig:Motivation}
\end{figure}
\section{Introduction}

The short video has become an indispensable medium for people to acquire knowledge and share experiences in their daily lives. However, an unedited video cost minutes or hours for a person to collect meaningful moments by going through the whole video. How to automatically identify the most interesting moments for specific users has recently drawn attention from the research community.

Unlike generic highlight detection methods \cite{yao2016highlight,gygli2016video2gif,xiong2019less}, we focus on personalized video highlight detection, where the user’s previously created historical highlight video segments are used as guidance to detect the interesting moments. To the best of our knowledge, only two methods are proposed to solve this problem. Molino and Gygli \cite{garcia2018phd} directly concatenate the history feature into the input video feature to perform segment level prediction by using a ranking model. Rochan et al. \cite{rochan2020adaptive} introduce a  temporal-adaptive instance normalization layer that encodes history for frame-level highlight predictions. Although promising results they achieved, there are still some limitations. Firstly, They regard the user’s history as holistic information to predict the user’s preference based on the assumption that there is only one preference in its history. However, it is somehow against common sense in the real world because the user’s interests are inherently diverse. Suppose a user prefers to travel, swimming, pets, et al. It is not suitable to extract a holistic preference to represent such diverse interests (Fig.~\ref{fig:Motivation}). Secondly, because there are many repeated or similar shots in a video while only a few positive samples are marked, some unlabeled samples are actually positive. However, the current methods treat all unlabeled samples as negative ones, leading to the wrong label assignment \cite{frenay2013classification}.

In this paper, we propose a preference reasoning framework (PR-Net) for frame-level personalized video highlight detection, which overcomes the above limitations. Intuitively, if a video frame is semantically similar to one of the user's history highlights, it could be regarded as an interesting moment. Based on this observation, we first attend the query frame to history frames to obtain attention weights and then forming the user-specific preference embedding by the attention-weighted sum of history frame embeddings. Considering the generic preference is also an important reference to judge the degree of highlight, especially when the user's history is missing. We set a learnable generic preference embedding, and it combines the user-specific preference embedding to generate the comprehensive preference embedding for more overall highlight detection. In the end, to ensure a more compact and differentiable metric space, a non-highlight preference is used to consider the degree of non-highlight for each query frame. We first infer the degree of highlight and non-highlight of each query frame based on the semantic similarity to its comprehensive preference embedding and non-highlight preference embedding, respectively. And then constraint the two similarities of all video frames by a new contrastive loss function. Note that we only train those most non-highlight frames as negative samples to alleviate the wrong label assignment problem.

We conduct the experiments on the PHD-GIF \cite{garcia2018phd} dataset, the only related large-scale dataset for this task. The results show that our method significantly outperforms the state-of-the-art methods, with the relative improvement of 12\% in mean accuracy precision.

The contributions of our framework are as follows.
\begin{itemize}
\item {We propose a novel preference reasoning framework named PR-Net for personalized video highlight detection.}
\item {We propose a novel bi-directional contrastive loss to train a more compact and differentiable metric space for frame-level content understanding.}
\item {Our method significantly outperforms state-of-the-art methods on a large-scale dataset.}
\end{itemize}

\section{Related Work}

Personalized video highlight detection aims to detect the interesting moments given a new video based on the previously highlighted video segments the user is interested in. The task is closed to video summarization to find all key events that represent the whole video. Our method is also related to the personalized recommendation if treating the video frame as the item to recommend.

\subsection{Video Highlight Detection}
Many prior methods \cite{jiao2017video,gygli2016video2gif,sun2014ranking,yao2016highlight,yu2018deep} have been proposed to solve generic video highlight detection. They typically learn a ranking model to measure the degree of the highlight of video segments without the user’s preference. Only two methods are predicting the user’s preference from history for personalized video highlight detection. Molino and Gygli \cite{garcia2018phd} apply a shot detection algorithm \cite{gygli2018ridiculously} for sampling a set of video segments. And then, directly concatenate the history feature and the video segment feature into a fused feature for highlight prediction. However, the usage of the shot boundary detector makes their pipeline computationally complex and expensive. Rochan et al. \cite{rochan2020adaptive} employ a temporal-adaptive instance normalization that explicitly considers the user’s history while generating highlights. These two methods share the same weakness when comparing our method. They regard the user’s history as holistic information to predict the user’s preference but negating the inherent diversity of the user’s interests. Besides, they train all unlabeled video frames as non-highlight samples, resulting in wrong label assignment because some unlabeled frames are semantic similar to highlight samples.

\begin{figure*}
  \centerline{\includegraphics[width=1\textwidth]{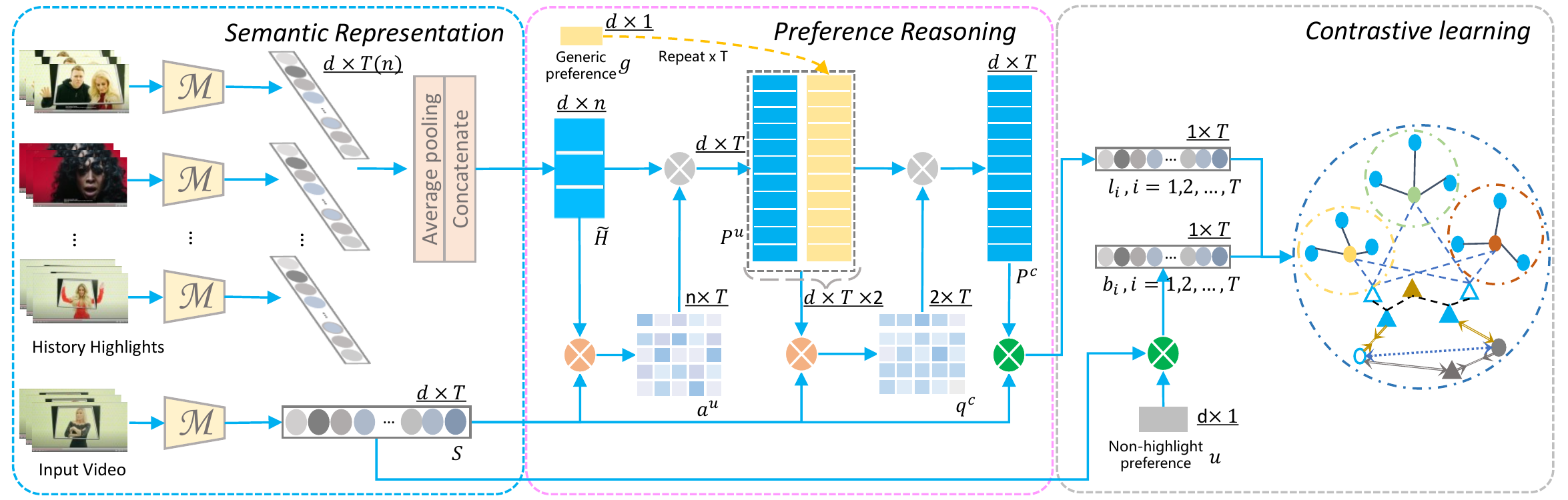}}
  \caption{Illustration of our framework. Firstly, each frame's context semantic representation in input video and history highlights are produced by an encoder $\mathcal{M}$ (left). Secondly, a stacked self-attention module entailing two stages of attention. It first attends to history embedding to form user-specific preference embeddings $P^{u}=\{p^{u}_{1},p^{u}_{2},...,p^{u}_{T}\}$, one for each input video frame. The second stage attends to a user-specific preference embedding $P^{u}$ and the generic preference embedding $g$ to form comprehensive preference embeddings $P^{c}=\{p^{c}_{1},p^{c}_{2},...,p^{c}_{T}\}$ for each of the input video frames (middle). Lastly, a bi-directional contrastive loss is proposed to constraint the degree of highlight $l_{i}$ and non-highlight $b_{i}$, which are similarities between each input frame embedding to its comprehensive preference embedding $P^{c}$ and non-highlight preference embedding $u$, respectively(right).}
  \label{fig:framework}
\end{figure*}

\subsection{Video Summarization}
Unlike video highlight detection that aims to find the most interesting moments, video summarization is to select a set of key events to represent the whole video semantically. Early approaches \cite{khosla2013large,kim2014reconstructing,lee2012discovering,lu2013story,mahasseni2017unsupervised,ngo2003automatic,panda2017collaborative,song2015tvsum,zhou2018deep,zhang2018retrospective,rochan2018video} select some keyframes from a video to satisfy the diversity and representativeness properties based on hand-crafted heuristics, which is not accurate and robust enough. Some learning-based method \cite{gong2014diverse,gygli2014creating,gygli2015video,rochan2018video,zhang2016summary,zhang2016video,zhao2017hierarchical,hong2020mini} directly learn from human-annotated training data that show superior performance. However, these are generic methods without the user’s preference involved. And it is impractical to train a particular model for every new user. There are also some personalized video summarization methods. They either use meta-data \cite{babaguchi2007learning,jaimes2002learning,takahashi2007user} or user textual query \cite{ulyanov2017improved,sharghi2016query,liu2015multi} to consider user’s preference. While we adopt a more convenient way that exploits visual features from the user’s history to predict the user preference.

\subsection{Personalized Recommendation}
Personalized recommendation systems are widely utilized in a variety of areas, including movies \cite{pecune2019model,gao2017collaborative}, music \cite{kim2019music,gao2017collaborative} and e-commerce \cite{xian2019reinforcement,xian2020cafe,wang2019explainable,cao2019unifying,gu2020hierarchical}. Several methods focus on incorporating knowledge graphs (KGs) into recommender systems for coupling recommendation \cite{pecune2019model,xian2020cafe,wang2019explainable,cao2019unifying}. Xian et al. \cite{xian2019reinforcement} utilize reinforcement learning and a policy-guided graph search algorithm to sample reasoning paths for the recommendation. Gao et al. \cite{gao2017collaborative} profile the users' dynamic preferences by a vector autoregressive model. Gu et al. \cite{gu2020hierarchical} model users' hierarchical real-time preferences via a pyramid recurrent neural network for e-commerce recommendation. In this paper, we explore a stacked-attention model for preference reasoning in video highlight detection.

\section{Method}
Given a video $V$ with $T$ frames, we aim at predicting the degree of highlight and non-highlight of each frame under the guidance of the user's history. Here the history $H=\{h_{1}, h_{2},..., h_{n}\}$ donates $n$ video segments the user is interested in. Note that the frame numbers vary in different videos, and the number of history video segments varies in different users.

Our framework is illustrated in Fig.~\ref{fig:framework}. Firstly, each frame's contextual semantics embedding in history and the input video are processed by an encoder $\mathcal{M}$. Secondly, each input video frame's comprehensive preference, represented as the weighted sum of history and generic preference embedding, is reasoned via a stacked self-attention module. Lastly, each input frame embedding's similarity to its comprehensive preference embedding and non-highlight preference embedding, respectively reflecting the degree of highlight and non-highlight, is constrained by a new proposed bi-directional contrastive loss. In the following, we present details and give more insights into our framework.

\subsection{Frame-level Context Semantic Representation}

To alleviate the computational burden, following prior work \cite{rochan2020adaptive}, we utilize a feature extractor that pre-trained on Sports 1M dataset \cite{karpathy2014large} to extract C3D \cite{tran2015learning} (conv5) layer features as the local feature representation of each frame in the input videos and user’s history. However, the C3D feature is fixed and only encodes local appearance and motion features during the training process. The contextual semantics of each frame is also important to estimate the interestingness. For example, a complete action or event is more interesting than the background and the shot boundary. Therefore, we design an encoder $\mathcal{M}$ to capture both the short and long-range features of each frame. 

\begin{figure}
  \centerline{\includegraphics[width=0.48\textwidth]{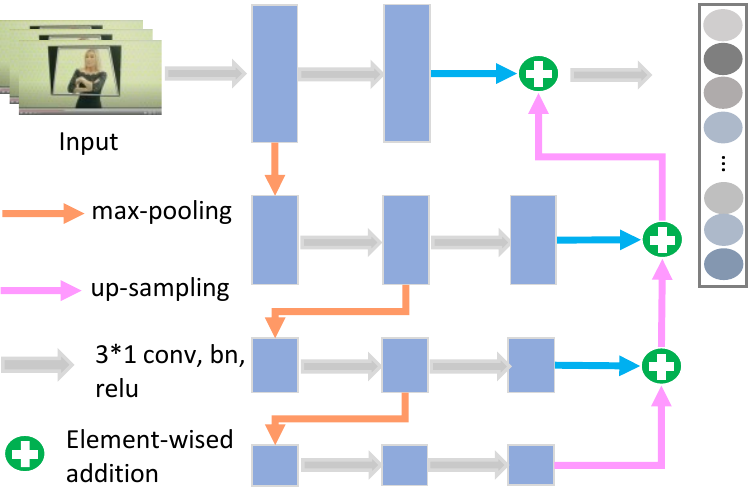}}
  \caption{The architecture of the encoder $\mathcal{M}$. The dimension of all convolutional layers is $d$, the same as the frame feature.}
  \label{fig:encoder}
\end{figure}

The encoder $\mathcal{M}$ is inspired by U-net \cite{ronneberger2015u}, the network structure is shown in Fig.~\ref{fig:encoder}. The input is a video with $T$ frames represented as C3D features $V\in\mathcal{R}^{d\times T}$, the output $S\in\mathcal{R}^{d\times T}$ is the contextual semantics of each frames in the video. Here $d$ indicates the feature dimension.
\begin{equation}\label{equ:embedding}
S = \mathcal{M}(V).
\end{equation}

\subsection{Attention-guided Preference Reasoning}

Considering the diversity of user's preferences is beneficial for an accurate recommendation system. Unfortunately, current highlight detection methods \cite{garcia2018phd,rochan2020adaptive} directly learn the holistic preferences from the user's history, which ignores the inherent diversity of the user's interests. We provide the preference reasoning module details on how to tackle this issue in the following.

Preference reasoning module expects two inputs: the contextual semantic embedding of the input video $S=\{s_1, s_2,...,s_T\}$; the contextual semantic embedding of the user's history $\widetilde{H}=\mathcal{M}(H)=\{\widetilde{h}_1,\widetilde{h}_2,...,\widetilde{h}_n\}$, where $\widetilde{h}_j,j=1,2,...n$ is the  representation for the user history by averaging the features of frames in the video segment. 

Considering that a video frame $s_{i}$ could be interesting if the contextual semantic is similar to one of the user’s histories. Therefore, we retrieve the user’s most relevant history highlight to obtain the user-specific preference embedding $p^{u}_i$ for this frame. 

\begin{equation}\label{equ:userPreference}
p^{u}_{i} = \sum^{n}_{j=1}\widetilde{h}_{j}*a^{u}_{ij},
\end{equation}
where $*$ is the multiplication operation, $a^{u}_{ij}$ is the attention weight and obtained as follows:
\begin{equation}\label{equ:userAttention}
a^{u}_{ij} = \frac{\exp(\lambda q^{u}_{ij})}{\sum^n_{j=1}\exp(\lambda q^{u}_{ij})} ,q^{u}_{ij} = \frac{s_{i} \cdot \widetilde{h}_{j}}{\|s_{i}\|\|\widetilde{h}_{j}\|}, \\
\end{equation}
where $\cdot$ is the dot product operation. $\lambda$ is the inversed temperature of the softmax function \cite{chorowski2015attention} and set to be 9 empirically.

\begin{figure}
  \centerline{\includegraphics[width=0.48\textwidth]{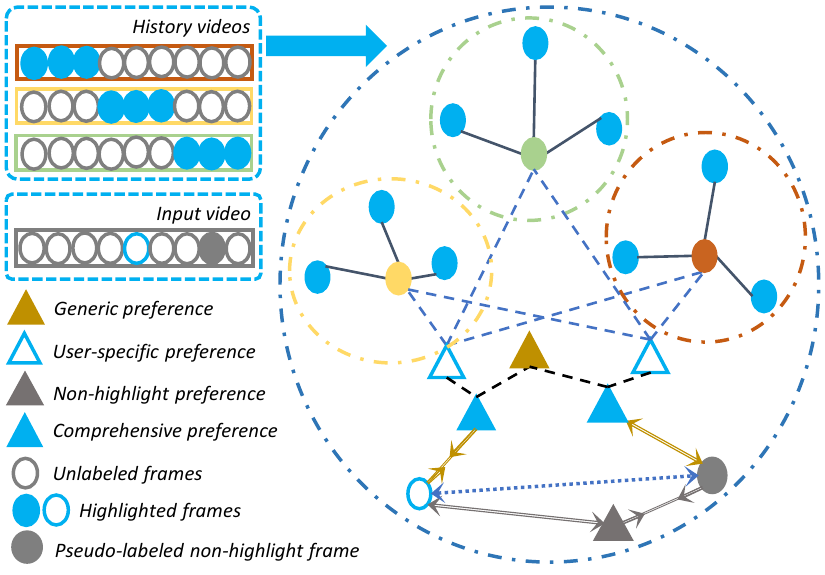}}
  \caption{Illustration of the bi-directional contrastive learning. The history embedding is the average feature of frames, represented as the yellow, green and red circle. The user-specific preference is the weighted sum of history embeddings (represented by the blue dotted lines). The attention weights are the similarity between the query frame and the history embeddings. The comprehensive preference is the weighted sum of the user-specific and generic preferences (represented by the black dotted lines). As described in section 3.3, we constraint the relationship of the above notations (represented by the solid lines).}
  \label{fig:contrastive_learning}
\end{figure}

Next, considering the generic preference is also an important reference to estimate the degree of highlight, especially when the user’s history is missing. Therefore, we take the user-specific preference embedding $p^{u}_{i}$ and the generic preference embedding $g$ to form the comprehensive preference embedding $p^{c}_{i}$ for more overall highlight detection.
\begin{equation}\label{equ:comprehensivePreference}
p^{c}_{i} =\frac{p^{u}_{i}*q^c_{1}}{q^c_{1}+q^c_{2}}+\frac{g*q^c_{2}}{q^c_{1}+q^c_{2}},
\end{equation}
where the generic preference embedding $g$ is learnable and is shared for all frames. The weight terms $q^c_{1}$ and $q^c_{2}$ are calculated by following:
\begin{equation}\label{equ:comprehensivePreference}
q^{c}_{1}=\exp(\frac{\lambda s_{i} \cdot p_{i}^{u}}{\|s_{i}\|\|p_{i}^{u}\|}),q^{c}_{2}=\exp(\frac{\lambda s_{i} \cdot g}{\|s_{i}\|\|g\|}).
\end{equation}

We obtain $T$ comprehensive preference embeddings $p^{c}_{i}, i=1,2,...,T$, one for each input frame. The degree of highlight $l_i$ for $i$-th frame is represented as the cosine similarity of the comprehensive preference embedding $p^{c}_{i}$ and its contextual semantic embedding $s_{i}$. Note that we omit the normalization term to reduce the computational burden.
\begin{equation}\label{equ:comprehensivePreference}
l_i = s_{i} \cdot p^{c}_{i}.
\end{equation}

We show a feasible explanation of the results. If the $i$-th frame is predicted to be interesting, we could check the attention weight $q^{c}_1$ and $q^{c}_2$ to see how generic preference and user-specific preference affect the prediction. Moreover, the attention weights $a^u_{ij}$ reflect the importance of history highlights for the $i$-th frame prediction.

\subsection{Bi-directional Contrastive Learning}

To ensure a compact and differentiable embedding space, we introduce a learnable non-highlight preference embedding $u$ to measure the degree of non-highlight $b_{i}$ for the $i$-th frame.
\begin{equation}\label{equ:comprehensivePreference}
b_i = s_{i} \cdot u.
\end{equation}

We constraint the relationship between positive frames and negative frames from two aspects. On the one hand, we set the comprehensive preference embeddings as the anchors, pull in the positive frame embeddings while pushing away those negative frame embeddings. On the other hand, we set the non-highlight embeddings as the anchors, pull in the negative frame embeddings while pushing away those positive frame embeddings (Fig.~\ref{fig:contrastive_learning}). The loss function is formulated as the following:
\begin{equation}\label{equ:contrastiveLoss}
\mathcal{L} = \sum_{y \in \Omega}\log( \widetilde{l}_{y} + \widetilde{b}_{y})+\sum_{x \in \mho}\log( \widetilde{l}_{x} + \widetilde{b}_{x})-(\sum_{y \in \Omega}\widetilde{l}_{y} + \sum_{x \in \mho}\widetilde{b}_{x}),
\end{equation}
where $\Omega$ is the set of frames with the positive label. $\mho$ is the set of unlabeled frames with the top $K$ highest value of $b_{i}$. $K$ is set to be five times the size of $\Omega$ empirically. The $\widetilde{l}_{i}$ and $\widetilde{b}_{i}$ is the relative degree of highlight and non-highlight among all frames, respective, calculated as follows.
\begin{equation}\label{equ:ranking}
\widetilde{l}_{i} = \frac{\exp(l_{i})}{\sum^T_{i=1}\exp(l_{i})}, \widetilde{b}_{i} = \frac{\exp(b_{i})}{\sum^T_{i=1}\exp(b_{i})}.
\end{equation}

\subsection{Implementation Details}
To employ multi-GPU training, we cut a video into fixed-length segments (256 frames) and only train those segments that contain highlight frames, and the batch size is set to be 32 (32 users). We take the whole video as input for testing, and the batch size is set to be 1. The entire framework is built on PyTorch 1.50. The training time is about 7 hours for 150 epochs on 8 NVIDIA Tesla V100 GPUs. It is optimized by the adam optimizer using the default configuration. We train our model from scratch except for the C3D feature extractor.

\begin{figure}
  \centerline{\includegraphics[width=0.5\textwidth]{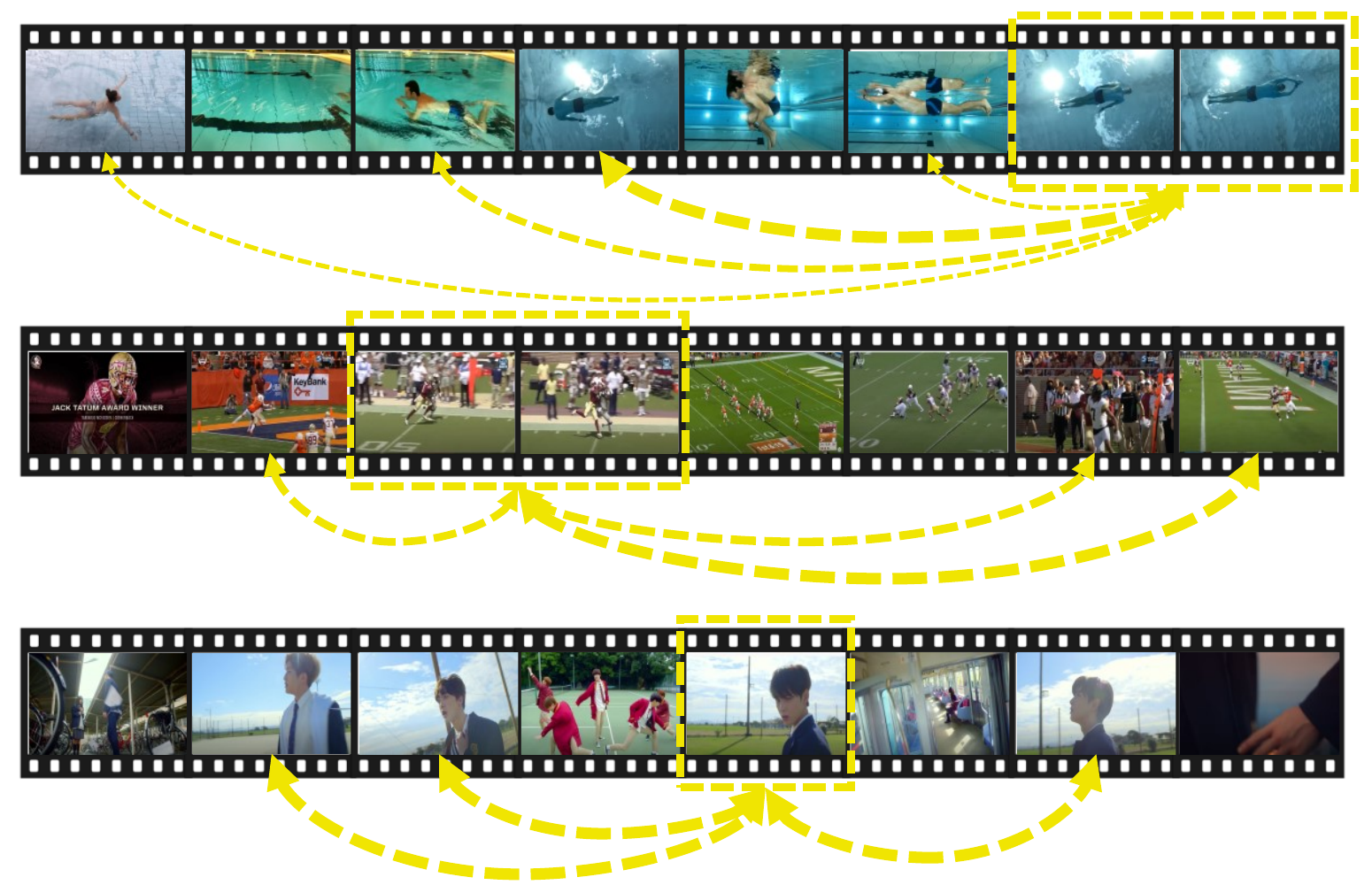}}
  \caption{Visualization of the semantically similar frame in the same video. The frames in the yellow box are ground truth highlights, while others are unlabeled frames. We use the width of the yellow line to indicate the similarity between the ground truth and the unlabeled frame. The three videos are swimming, football, and Songs, respectively, show that a video contains similar frames is a common phenomenon in this dataset.}
  \label{fig:repeat_frame}
\end{figure}

\begin{figure*}
  \centerline{\includegraphics[width=1\textwidth]{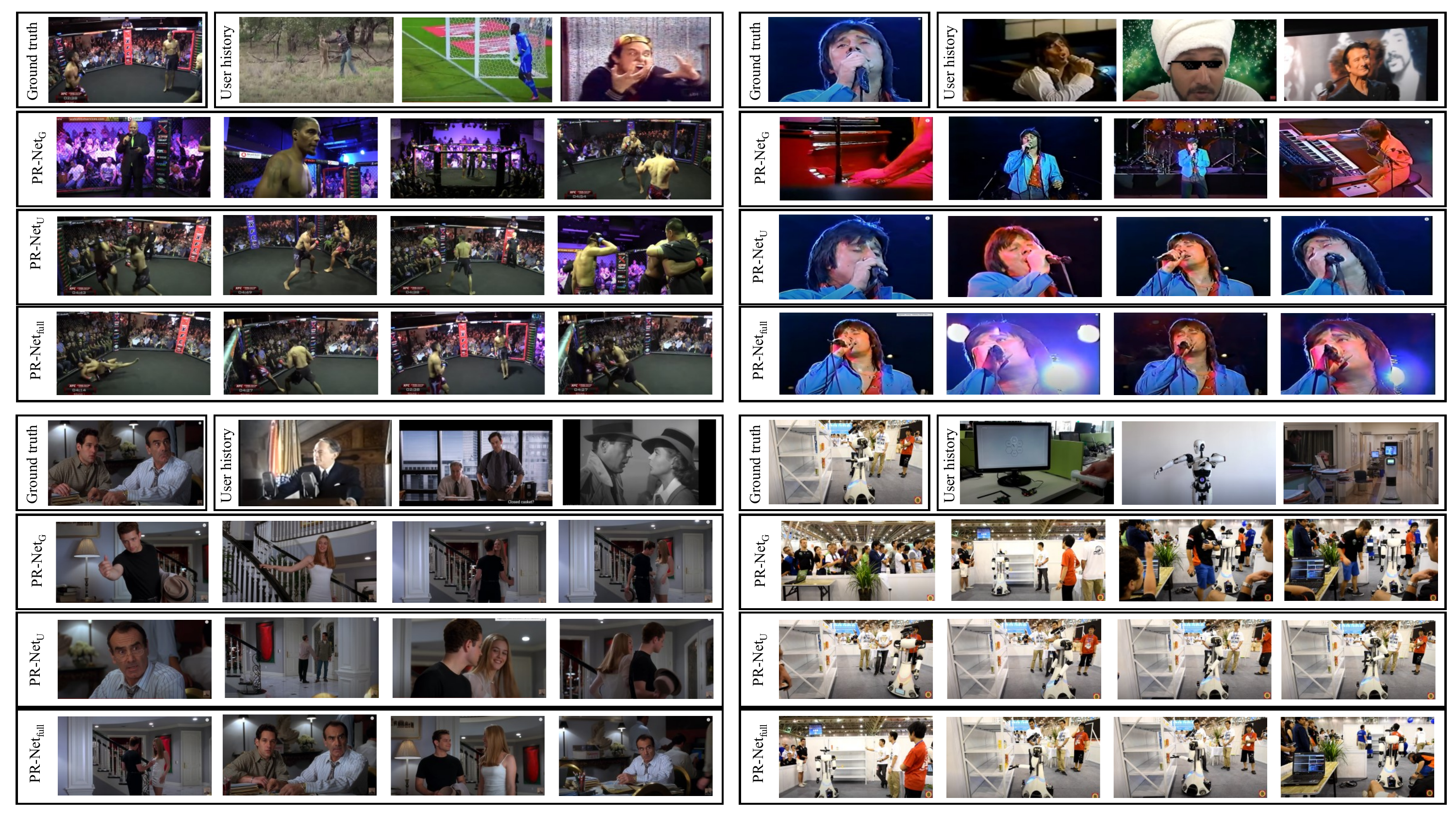}}
  \caption{Qualitative evaluation of different configurations of our method. We show examples of our method with only generic preference (PR-Net$_G$), with only user-specific preference (PR-Net$_U$) and our full method (PR-Net$_{full}$) on four videos. In addition, we show the user’s history (multiple GIFs) and a few sampled frames from the highlight predictions of the three models for each video. These visualizations indicate that the user-specific preference well captures the user’s diverse interests from history, resulting in more reasonable predictions than only generic preference.}
  \label{fig:visualization}
\end{figure*}

\section{Experiments}
In this section, the experiments conducted on a large-scale video highlight detection dataset show that our method significantly outperforms current state-of-the-art methods. We also discuss its advantages and limitations. In the end, we conduct ablation experiments to highlight the contributions of each module and present their insights.
\subsection{Experiments Configuration}
\paragraph{Dataset}
PHD-GIFs \cite{garcia2018phd} provides a large-scale dataset that contains the user's history information for video highlight detection. The released dataset consists of 11,972 users for training, 1000 users for validation, and 850 users for testing. Note that there is no overlap among users in these three subsets. For training, each user has at most twenty annotated videos, with one or more segments marked as the highlight in each video. In summary, there are 222,015 annotations in 119,938 videos.

The dataset has some notable properties. First, the annotations cover various topics, such as cartoons, pets, music videos and beautiful scenes. Moreover, the user has consistent interests in one or more topics, which provide reliable guidance for personalized video highlight detection. Secondly, only a small proportion of frames are marked as the highlights in the dataset. It is about 1:50 for positive vs unlabeled. Lastly, we observe that there are many similar shots in a video, but only one of them is annotated as the highlight (Fig~\ref{fig:repeat_frame}). The incomplete annotation may cause ambiguity when training the unlabeled frames as negative samples.

Only the YouTube video ID of videos are provided in the dataset for download from the web. Because some video links are no longer available, we only download 95,111 videos with a total duration of 10,212 hours. Following Rochan et al. \cite{rochan2020adaptive}, we extract the frame-level C3D feature from all videos by a pre-trained model. The processing time is about two weeks on 8 NVIDIA Tesla V100 GPUs. We process 9478 users for training, 750 users for validation and 675 users for testing in the end. Since the data processing is tedious, time-consuming and computationally expensive, we are willing to release the C3D features of the whole dataset to the community for future works.

\paragraph{Evaluation Metric}The mean Average Precision (mAP) is adopted to measure the performance of our method, which is also commonly used by previous works \cite{yao2016highlight,rochan2020adaptive}. Follow Rochan et al. \cite{rochan2020adaptive} we treat videos separately when calculating the mAP.

\paragraph{Baselines}Our method is compared with other state-of-the-art methods on the dataset, including FCSN \cite{rochan2018video}, Video2GIF \cite{gygli2016video2gif}, PHD-GIFs \cite{garcia2018phd} and A-VHD \cite{rochan2020adaptive}. FCSN is the state-of-the-art video summarization method, which is generic video highlight detection for comparison. Video2GIF is a state-of-the-art model for video highlight detection. Following Rochan et al. \cite{rochan2020adaptive}, we use the publicly available pre-trained model for comparison. PHD-GIFs is a state-of-the-art personalized video highlight detection method. We aggregate frames into the five-second shots in every video to report the performance for a fair comparison. A-VHD is also a state-of-the-art method for personalized video highlight detection.

\begin{figure*}
  \centerline{\includegraphics[width=1\textwidth]{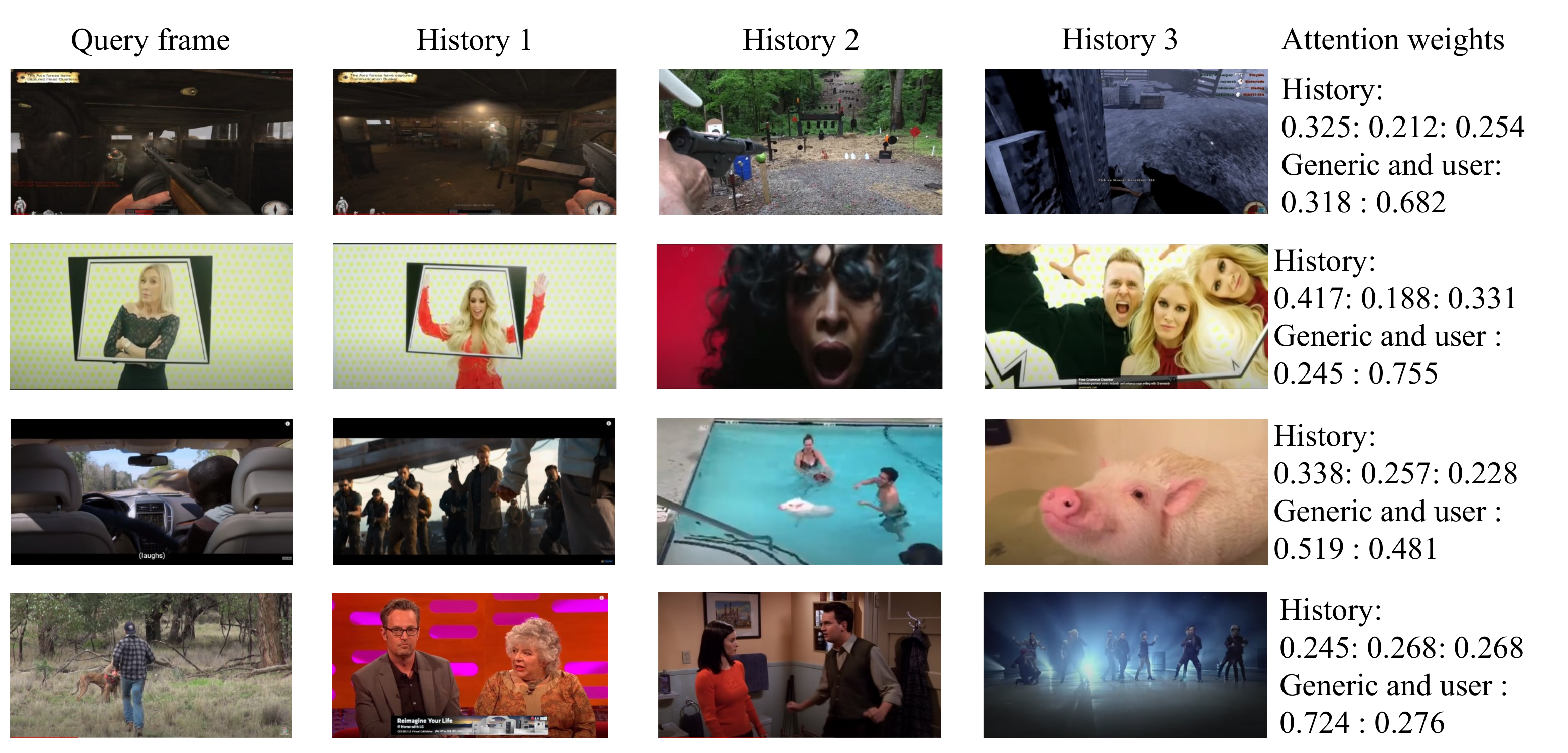}}
  \caption{Attention visualization for the detection results. We select four cases under different scenarios to present three types of attention weights (for history embeddings, user-specific preference embedding and generic preference embedding). Each row stands for a case. Note that we only show the top three history highlights for each case, and the remaining histories are not meaningful than the top 3. In case 1 (the first row), all history highlights are semantically related to the query frame. Case 2 presents a highly subjective scenario in which the semantics are ambitious in the history and query frame. Case 3 and Case 4 show that the user's interests are diverse. However, there is no relevant history with the query frame in case 4.}
  \label{fig:genericAndUser}
\end{figure*}

\begin{table}[h]
	\centering
	\caption{Comparison with state-of-the-art methods on the PHD-GIFs dataset. The performance of Video2GIF, FCSN$^*$ and A-VHD$^*$ are reported in \cite{rochan2020adaptive}. While others are evaluated on our dataset. PR-Net is our full method. }\label{tab:comparisons}
	\scalebox{1}{
	\begin{tabular}{c c}
        \hline
		\multirow{2}*{Method} &
		\multirow{2}*{mAP(\%)}\\

		~ & ~ \\
		\hline
		Video2GIF \cite{gygli2016video2gif}& 14.75 \\
		FCSN$^*$ \cite{rochan2018video}& 15.22 \\
		A-VHD$^*$ \cite{rochan2020adaptive}& 16.73 \\
        \hline
        FCSN \cite{rochan2018video}& 15.15 \\
		PHD-GIFs \cite{garcia2018phd}& 16.25 \\
		A-VHD \cite{rochan2020adaptive}& 16.68 \\
        PR-Net  & \textbf{18.66} \\ 
		
	\end{tabular}}
\end{table}

\subsection{Results and Discussions}
We quantitatively compare our method with other state-of-the-art methods. Moreover, its explainability, advantages and limitations are presented in the following.
\paragraph{Comparison results}
The results show that our method significantly outperforms current state-of-the-art methods (Table~\ref{tab:comparisons}). Compared with two generic video highlight detection methods (Video2GIF and FCSN) without using the user's history, our method efficiently utilizes the user's preference for guiding the highlight detection and achieves much more accurate results. The mAP is relatively improved by 28.24\% and 23.17\%, respectively. PHD-GIFs directly concatenates the average history feature into the input video feature to perform segment-level prediction. A-VHD predicts the affine parameters of the adaptive instance normalization from the user's history, where the predicted affine parameters can be regarded as the user's preference for guiding personalized video highlight detection. However, the limitation of the two methods is that the user preference is determined by the holistic history highlights, implying only one interest in the user's history. Therefore, they may not be suitable solutions when user's interests are diverse. While in our method, we overcome the limitation by reasoning the most relevant history highlight for each input frame, results in more accurate detection results, with the relative improvement of 12\%.

\paragraph{Attention Visualization}

The personalized video highlight detection is subjective and highly depends on the user's preference. Intuitively, the query video frame could be regarded as a highlight if it is semantically similar to one of the user's highlight histories. Our method utilizes the attention mechanism to find the most relevant highlight history for the query video frame. After checking many cases, we discovered that the learned attention weights almost reflect the semantic similarity between the query and history frames, which indicates a potential way to explain the detection results. As shown in Fig.~\ref{fig:genericAndUser}, we visualize the results with the top three most relevant history highlights of four cases. 

In case 1 of the query frame, it is shooting an enemy with a machine gun. Observing the attention weight, we find the history with the highest attention weight is most semantically similar to the query frame, shooting an enemy with a pistol. For history 2 and 3, shooting without the enemy, the attention weight is lower than history 1. 

Case 2 is highly subjective, and the semantics is ambitious. It seems that all history highlights are relevant to the query frame. However, we can observe that the most relevant history share the same background with the query frame. 

The user's interests are diverse in case 3. It seems that the user prefers a movie, swimming and pets. We can see that the attention weight reflects semantic similarities. Moreover, the generic weight is raised compared with case 1 and case 2 because the query frame is not highly relevant to the historical highlights.

There are no relevant history highlights in case 4 from our view. In this case, the generic preference significantly affects the degree of the query frame's highlight. 

\paragraph{Limitations and Future Work}
The detection results of our method are sometimes confusing, which is hard to tell it is interesting or not. We think the subjectiveness of the user annotation mainly causes it as the model is directly learning from the training data. Further enhancing the semantic representation by multiple feature extractors may be a promising way to solve the problem. We left it in future work.

\begin{table}[h]
	\centering
	\caption{Ablation Study Experiments. PR-Net(*) stand for the different modification based on our method.  (Ablation experiments section). PR-Net$_{full}$ is our full method.}\label{tab:ablation}
	\scalebox{1}{
	\begin{tabular}{c c c}
        \hline
        \multirow{2}*{Ablation target} &
		\multirow{2}*{Method} &
		\multirow{2}*{mAP(\%)} \\

		~ & ~ \\
	    \hline
	    Representation & PR-Net$_{residual}$ & 16.69 \\
        \hline
		\multirow{3}{*}{Loss Strategy} & PR-Net$_{ranking}$ & 16.85 \\
		 & PR-Net$_{CE}$ & 17.09 \\
		 & PR-Net$_{triplet}$ & 17.26 \\
		\hline
		\multirow{4}{*}{Attention Strategy} & PR-Net$_G$ & 16.78 \\
		& PR-Net$_U$ & 17.57 \\
		& PR-Net$_{meanU}$ & 16.87 \\
		& PR-Net$_{meanU+G}$ & 17.21 \\
		\hline
		
		\multirow{2}{*}{History Size}
		
		& PR-Net$_{h0}$ & 16.78 \\
		& PR-Net$_{h5}$ & 17.48 \\
		\hline
		- & PR-Net & \textbf{18.66} \\ 
		
	\end{tabular}}
\end{table}

\subsection{Ablation Experiments}
We conduct extensive experiments to highlight the effectiveness of different modules in our framework.

\textbf{PR-Net$\bm{_{residual}}$:} Replacing the encoder $\mathcal{M}$ with a residual block, in which the number of the parameters is larger than $\mathcal{M}$. The residual block is applied to the C3D feature of each frame. 

\textbf{PR-Net$\bm{_{ranking}}$, PR-Net$\bm{_{CE}}$ and PR-Net$\bm{_{triplet}}$:} We respectively replace the bi-directional contrastive loss with the ranking loss (PR-Net$_{ranking}$) and the Cross-Entropy loss (PR-Net$_{CE}$) as they are also used in other highlight detection method \cite{yao2016highlight,rochan2020adaptive}. Note that for the implementation of PR-Net$_{ranking}$, we abandon the non-highlight preference and only measure each frame's degree of highlight. We also present the performance when using the triplet loss \cite{cheng2016person} (PR-Net$\bm{_{triplet}}$).

\textbf{PR-Net$\bm{_{G}}$} and \textbf{PR-Net$\bm{_{meanU}}$:} To evaluate the effect of user-specific preference, two variants of the PR-Net are studied: Directly removing the user-specific preference (PR-Net$\bm{_{G}}$), using the mean history embeddings to replace the user-specific preference (PR-Net$\bm{_{meanU}}$).

\textbf{PR-Net$\bm{_{U}}$} and \textbf{PR-Net$\bm{_{meanU+G}}$:} For the evaluation of the generic preference, we explore the two variants of the PR-Net: without generic preference (PR-Net$\bm{_{U}}$), adding generic preference on PR-Net$\bm{_{meanU}}$ variant (PR-Net$\bm{_{meanU+G}}$).

\textbf{PR-Net$\bm{_{h0}}$} and \textbf{PR-Net$\bm{_{h5}}$:}  Two experiments are conducted to evaluate the effect of history: no history information involved (PR-Net$\bm{_{h0}}$) and restrict the history size with five PR-Net$\bm{_{h5}}$.









\vspace{-1.5ex}

\paragraph{Effect of frame-level contextual  semantics representation}
To verify how contextual semantic representation affects the performance, we only consider the short-range features of each frame (PR-Net$_{residual}$). Compared with our full method (PR-Net$_{full}$), the performance is dramatically decreased (from 18.66\% to 16.69\%). The result shows that contextual semantic representation is vital for highlight detection.
\vspace{-1.5ex}

\paragraph{Effect of bi-directional contrastive loss}
Compared with PR-Net$_{ranking}$, PR-Net$_{CE}$ and PR-Net$_{triplet}$ that respectively use ranking loss, Cross entropy loss and triplet loss, our carefully designed bi-directional contrastive loss provide more reasonable constraints in the metric space for video highlight detection, resulting in better performance.

\vspace{-1ex}

\paragraph{Effect of generic preference}
We remove the generic preference embedding from our method and only utilize the user-specific preference to measure the degree of highlight (PR-Net$_U$, and PR-Net$_{meanU}$). Note that they are already outperforming the the-state-of-the-art method. However, because they do not consider the generic preference, especially when the relevant history is missing, the performance is worse than that with generic preference(PR-Net$_{meanU+G}$ and PR-Net$_{full}$)

\paragraph{Effect of user-specific preference}
It becomes the generic highlight detection model if without user-specific preference (PR-Net$_G$). As shown in Fig.~\ref{fig:visualization}, the detection results of using user-specific preference are obviously closer to history than those without it. Experiment results show that frame-level user-specific preference is vital for boosting performance and lifting the mAP from 16.78\% to 18.66\%. Compared with PR-Net$_{meanU}$ that adopt mean history embeddings as the user-specific preference (holistic preference), the usage of attention mechanism to produce user-specific preference is functional, which implies that our method offers a more efficient way to utilize the user's history for personalized video highlight detection.

\paragraph{Effect of history size}
Theoretically, the performance of our method is benefits from more history highlights for reference. Therefore, we also show the performances when restrict the maximal size of history to be zero (PR-Net$_{h0}$) and five (PR-Net$_{h5}$), respectively, in which the performance is to meet our expectation.

\section{Conclusion}
In this paper, we explore a new paradigm incorporating the user's history for personalized video highlight detection. Unlike prior methods to predict a user preference from holistic history information, our attention-guided preference reasoning takes the diversity of user's interests into account to predict frame-level preferences. Experiments conducted on the large-scale dataset show that our method is much more accurate than state-of-the-art methods.

{\small
\bibliographystyle{ieee}
\bibliography{egbib}
}

\end{document}